\documentclass{article}

\usepackage[preprint]{neurips_2026}
\usepackage[utf8]{inputenc}
\usepackage[T1]{fontenc}
\usepackage{hyperref}
\usepackage{url}
\usepackage{booktabs}
\usepackage{amsfonts}
\usepackage{nicefrac}
\usepackage{microtype}
\usepackage{xcolor}

\usepackage{graphicx}
\usepackage[strings]{underscore}
\usepackage{grffile}
\usepackage{amsmath}
\usepackage{amssymb}
\usepackage{amsthm}
\usepackage{multirow}
\usepackage{wrapfig}
\usepackage{makecell}

\title{EGA: Adapting Frozen Encoders for Vector Search with Bounded Out-of-Distribution Degradation}

\author{%
  Dongfang Zhao \\
  Tacoma School of Engineering and Technology\\
  University of Washington\\
  \texttt{dzhao@uw.edu} \\
}

\begin{document}

\maketitle

\begin{abstract}
Vector search systems built on frozen vision encoders face queries from unseen classes at deployment, yet existing adapter training collapses under this shift: high-capacity adapters with global contrastive losses silently reassign unseen-class samples to wrong seen-class clusters, dropping worst-case Label Precision by over 40 points below the frozen baseline in our tests.
We propose Euclidean Geodesic Alignment (EGA), a residual adapter that couples three principles: zero initialization, local triplet loss, and hypersphere projection.
These collectively induce a self-limiting dynamic: triplets that already satisfy a small margin stop producing gradients, so the adapter automatically stops updating where the local geometry is already correct.
Our experiments show that at convergence $96.5\%$ of triplets are gradient-free, leaving unseen-class regions largely untouched while still enabling full-capacity refinement of seen classes.
Across five diverse out-of-distribution (OOD) benchmarks, EGA achieves the highest worst-case Label Precision on the four primary splits and a consistent improvement on the fifth.
The design also transfers to stronger backbones in addition to CLIP, and we provide an analytical justification linking gradient sparsity to bounded OOD perturbation.
\end{abstract}

\section{Introduction}

A vector similarity search system~\cite{johnson2019billion,subramanya2019diskann,chen2021spann,osdi25pipeann} deployed in production rarely operates on a fixed label space:
A retailer indexes product images today and adds a new category line next month; a content platform builds an index over existing tags and continually ingests media from emerging topics; a scientific image database is annotated for known phenotypes and queried for ones discovered later.
In such scenarios, the index and the query stream together cover a label space strictly larger than what was available when any model was trained or adapted.

The deployment of unseen classes sits awkwardly with how adapters over frozen encoders~\cite{clip_icml21,oquab2024dinov,10377550} are trained.
The dominant paradigm takes a labeled subset, optimizes the embedding geometry to separate classes, and ships the resulting adapter as a drop-in replacement for the frozen encoder's output.
The implicit assumption is that geometry tightened for seen classes will leave unseen-class geometry unchanged.
We show this assumption does not always hold: high-capacity adapters trained with global contrastive objectives reshape the entire embedding distribution, pulling unseen-class samples toward whichever seen-class clusters they happen to resemble in the pre-trained metric.
The retrieval system then inherits this distortion as the approximate nearest neighbor (ANN) search remains well-behaved geometrically while the neighbors may belong to the wrong semantic class.

The downstream retrieval system usually cannot know in advance which out-of-distribution (OOD) will dominate the deployed query stream.
The relevant criterion is therefore the \emph{worst-case Label Precision} across plausible OODs, not optimality on any single one.
Two design choices in current adapter training degrade this worst-case quantity.
First, the loss function: global contrastive objectives such as ICon~\cite{alshammariunifying} and SRL~\cite{dong2025improve} compute gradients over the full pairwise similarity matrix in every step, so seen-class supervision exerts pressure across the entire embedding distribution rather than only on labeled regions.
Second, the adapter capacity: a sufficiently expressive adapter has the representational room to comply with that pressure, and zero-initialized residual designs do not by themselves prevent the cumulative perturbation from reaching unseen-class regions.
Existing alternatives address one factor at a time.
Low-rank adapters such as LoRA~\cite{hu2022lora} cap the second factor, which avoids the worst OOD failures but limits in-distribution (ID) refinement, while keeping the global loss leaves the first factor untouched.
We argue that adapters intended to operate under unseen-class queries should be designed against both factors in a coordinated manner.

This work presents Euclidean Geodesic Alignment (EGA), a residual adapter designed against both factors of loss function and adapter capacity, simultaneously.
EGA pairs a high-capacity zero-initialized residual architecture with a small-margin triplet loss on the class-aware neighborhood graph of pre-trained embeddings, with $\ell_2$ projection onto the unit hypersphere.
Despite the technical intricacies, the underlying idea is straightforward, as follows.
On the one hand, the triplet loss induces a self-limiting training dynamic: as the adapter converges on seen classes, the fraction of margin-violating triplets decays rapidly.
On the other hand, the capacity to refine seen-class neighborhoods is preserved, but the loss stops applying pressure once seen-class neighborhoods are locally separated.
As a result, the global reshaping that destroys unseen-class geometry never accumulates.

We evaluate EGA on CIFAR-100 and ImageNet-1K (in-distribution) and five OOD benchmarks (CIFAR-10, FGVC-Aircraft, Food-101, ImageNet-1K held-out classes, and Oxford-IIIT Pet), comparing against the frozen CLIP baseline~\cite{clip_icml21}, global contrastive adapters (ICon~\cite{alshammariunifying}, SRL~\cite{dong2025improve}), and LoRA~\cite{hu2022lora} variants (InfoNCE~\cite{oord2018representation} and Triplet, both at $r=128$).
Across the four primary OOD benchmarks, EGA achieves the highest worst-case Label Precision (0.611), $4.2$ percentage points above the next-best method (LoRA+Triplet at 0.569), and is the only method whose Label Precision exceeds $0.6$ on every one of them; 
on Oxford-IIIT Pet, where the frozen baseline already reaches $0.9595$, EGA further raises LP@1 to $0.9646$ while achieving the highest ANNS Recall among all evaluated methods.
ICon and SRL collapse to $0.470$ and $0.375$ in their worst case (Aircraft); LoRA+InfoNCE to $0.538$ (Aircraft).
The benefits transfer to more recent backbones in addition to CLIP, with EGA improving Label Precision@1 by $4.75$ and $8.55$ percentage points on DINOv2-large~\cite{oquab2024dinov} and SigLIP~\cite{10377550}.

In summary, this work makes the following contributions:
\begin{itemize}
	\item We frame adapter tuning over frozen embedding encoders
	      from a perspective of downstream vector search systems, in which the index and
	      query stream together cover a label space larger than the labeled
	      subset available for adapter training, and we identify worst-case OOD
	      Label Precision as the relevant criterion under deployment uncertainty.

	\item We propose Euclidean Geodesic Alignment (EGA), a residual adapter design
	      that couples zero-initialized residual architecture, local triplet supervision,
	      and hypersphere projection.
	      This combination induces a self-limiting training dynamic: as seen-class neighborhoods become locally separated, the fraction of active triplets decays rapidly, empirically reaching $96.5\%$ zero-gradient triplets at steady state.
	      We also provide an analytical justification for this dynamic, showing that the unseen-class perturbation is bounded by the product of the active triplet ratio and the adapter's Lipschitz constant, and we explicitly state the conditions under which this bound is tighter than for global contrastive methods (Appendix~\ref{sec:appendix_theory}).

	\item Across five OOD benchmarks and three backbones (CLIP ViT-B/32, DINOv2-large, SigLIP), 
			EGA achieves the highest worst-case OOD Label Precision (0.611) on the four primary splits,
	      exceeding all five baselines by between $4.2$ and $23.6$ percentage points. On the fifth benchmark (Oxford-IIIT Pet), EGA further raises the already high frozen baseline from $0.9595$ to $0.9646$ while achieving the highest ANNS Recall among all evaluated methods.
\end{itemize}

\section{Background and Related Work}
\label{sec:lit}

Foundation models such as CLIP~\cite{clip_icml21} produce embedding spaces that encode both seen-class semantics and broader priors over unseen visual concepts.
Related work has documented degradation phenomena in adapted representations~\cite{lanza2026degradationfeaturespacecontinual, xue2025protocol}, and recent work documents analogous degradation in contrastive representations, including anisotropic collapse~\cite{zhao-etal-2024-representation, gao-etal-2021-simcse}, modality gap fracturing~\cite{10656738, eslami2025mitigate}, and loss of compositional structure~\cite{pal2025compositional, savietto2026geometryrepresentationalfailuresvision, 11093676}.
Recent strategies counter these issues with geometric constraints such as orthogonality regularization~\cite{ricci2026boldsymbollambdaorthogonality}.
\emph{EGA exploits the gradient sparsity of small-margin triplet losses
	to keep adapter updates concentrated on seen-class neighborhoods,
	empirically leaving unseen-class regions of the pretrained geometry
	largely intact.}

Recent representation learning frequently relies on global contrastive objectives to optimize the latent space~\cite{sui2026unicon, koromilas2025principledframeworkmultiviewcontrastive}.
A common paradigm jointly enforces alignment for positive pairs and uniformity across the feature distribution~\cite{10.5555/3524938.3525859}, applied in graph encoders~\cite{10.1609/aaai.v38i14.29479} and multimodal recommendation~\cite{zhou2025cm3calibratingmultimodalrecommendation}.
While uniformity improves robustness to noise~\cite{wu2026pca}, it treats all latent regions equally.
Other structural regularizers include unifying frameworks for multi-view alignment~\cite{alshammariunifying}, geometric constraints for imbalanced regression~\cite{dong2025improve}, decoupled objectives~\cite{kim2025decoupledcontrastivelearningfederated}, information-theoretic density control~\cite{Zhang_2025_ICCV}, and region-based distillation~\cite{jing2024fineclip}.
\emph{EGA replaces dense global optimization with sparse local triplet
	updates, which empirically reduces unseen-class distortion.}

Parameter-efficient adaptation is an approach for transferring large foundation models without full fine-tuning~\cite{pmlr-v97-houlsby19a, hu2022lora}, which dominates vision-language research~\cite{lin2025visionlanguagemodelssurvey}.
Lightweight methods, e.g., Visual Prompt Tuning~\cite{10.1007/978-3-031-19827-4_41} and Tip Adapter~\cite{tipadapter_eccv22}, specialize the model without modifying the frozen backbone and rival full fine-tuning on many tasks~\cite{sharma2025benchmarking}, while zero-initialization strategies further constrain the magnitude of pretrained-manifold perturbation~\cite{10377881}.
\emph{EGA admits a zero-initialized
	residual adapter, localized triplet objective, and parameter
	efficiency with gradient-sparse updates.}

Vector search is the foundational engine for scalable high-dimensional retrieval.
The systems community has developed indexing algorithms for billion-scale datasets, including FAISS~\cite{johnson2019billion}, HNSW~\cite{malkov2018efficient}, DiskANN~\cite{subramanya2019diskann}, and SPANN~\cite{chen2021spann}.
Recent work refines graph topologies~\cite{NEURIPS2024_bab1486c, NEURIPS2024_6dc63b40}, optimizes SSD-based search~\cite{osdi25pipeann, fast26odinann}, and accelerates distance computations~\cite{jaasaari2024lorann}, yet popular ANN implementations still suffer worst-case limitations~\cite{NEURIPS2023_d0ac28b7}.
\emph{EGA recalibrates the
	metric geometry before index construction, rendering search
	backends achieve higher semantic recall without algorithmic changes.}

\section{Methodology}
\label{sec:method}

\subsection{Problem Formulation}
\label{sec:formulation}

Given a frozen encoder $\phi : \mathcal{X} \to \mathbb{S}^{d-1}$ and labeled images $\mathcal{D}_\text{train} = \{(x_i, y_i)\}_{i=1}^{N}$ from \emph{seen} classes $\mathcal{Y}_\text{seen}$, we aim to learn an adapter $f_\theta : \mathbb{S}^{d-1} \to \mathbb{S}^{d-1}$ that supports accurate ANN retrieval on \emph{both} seen and unseen classes ($\mathcal{Y}_\text{unseen} \cap \mathcal{Y}_\text{seen} = \emptyset$), without access to unseen-class labels at training time.

We evaluate retrieval quality via \emph{Label Precision@$K$} (LP@K, semantic quality) and \emph{ANNS Recall@$K$} (AR@K, geometric fidelity):
\begin{equation}
    \text{LP}@K = \frac{1}{|\mathcal{Q}|} \sum_{q \in \mathcal{Q}}
    \frac{1}{K} \sum_{k=1}^{K}
    \mathbf{1}\!\left[y_{\hat{n}_k(q)} = y_q\right],
    \qquad
    \text{AR}@K = \frac{1}{|\mathcal{Q}|} \sum_{q \in \mathcal{Q}}
    \frac{\left|\mathcal{N}_K^\text{exact}(q) \cap \mathcal{N}_K^\text{approx}(q)\right|}{K},
    \label{eq:lp_ar}
\end{equation}
where $\mathcal{Q}$ is the query set, $\hat{n}_k(q)$ is the $k$-th approximate neighbor of $q$, and $\mathcal{N}_K^\text{exact}(q)$, $\mathcal{N}_K^\text{approx}(q)$ denote the exact and approximate top-$K$ neighbor sets. LP@K and AR@K capture distinct failure modes: tightening within-class clusters (semantic) vs.\ improving index fidelity (geometric).

\subsection{Manifold-Preserving Adapter Design}
\label{sec:design}

We design EGA to satisfy a single objective: \emph{refine seen-class neighborhoods while preserving the pretrained geometry wherever local separation is already sufficient}.
This translates into three design constraints, each addressed by a specific architectural choice.

\textbf{Constraint 1: Training must begin from the pretrained geometry.}
We enforce this via \emph{zero-initialized residual connections}.
Each EGA block applies a residual update:
\begin{equation}
    \mathbf{z}' = \mathbf{z} + g_\theta(\mathbf{z}),
    \label{eq:residual}
\end{equation}
where the final linear layer of $g_\theta$ is initialized to zero, making $f_\theta$ an identity map at initialization.
The adapter can therefore only \emph{refine} the frozen representation, never replace it.

\textbf{Constraint 2: Gradient pressure must be local and self-limiting.}
We replace global contrastive losses with a \emph{small-margin triplet loss} (Section~\ref{sec:training}) whose hinge formulation stops gradient flow once the anchor-positive distance falls below the anchor-negative distance by a margin $m$, making the training dynamic self-limiting.

\textbf{Constraint 3: Transformations must remain within the original metric space.}
We enforce this via $\ell_2$ \emph{projection back onto the unit hypersphere}:
\begin{equation}
    \hat{\mathbf{z}} = \frac{f_\theta(\mathbf{z})}{\|f_\theta(\mathbf{z})\|_2},
    \label{eq:l2norm}
\end{equation}
ensuring Euclidean distance and cosine similarity remain interchangeable.

Inside each residual block, a \emph{feature gating} module selectively amplifies retrieval-relevant dimensions through a bottleneck with ratio $1/4$:
\begin{equation}
    \text{Gate}(\mathbf{h}) = \mathbf{h} \odot \sigma\!\left(W_2\, \text{GELU}(W_1 \mathbf{h})\right),
    \label{eq:gate}
\end{equation}
where $W_1 \in \mathbb{R}^{(d/4) \times d}$, $W_2 \in \mathbb{R}^{d \times (d/4)}$, $\sigma$ is the Sigmoid function, and $\odot$ denotes elementwise multiplication, analogous to Squeeze-and-Excitation networks~\cite{hu2018squeeze}.

The complete forward pass chains two gated residual blocks followed by a final linear projection and $\ell_2$ normalization:
\begin{equation}
    \hat{\mathbf{z}} = \ell_2\!\left(\text{Refine}\!\left(\text{Block}_2\!\left(\text{Block}_1(\mathbf{z})\right)\right)\right).
    \label{eq:forward}
\end{equation}
The total trainable parameters (${\sim}4.2M$ for $d{=}512$, hidden dim 2048) are negligible compared to the frozen CLIP backbone (${\sim}88M$).

\subsection{Local Triplet Training}
\label{sec:training}

Given a mini-batch $\mathcal{B}$ sampled from $\mathcal{D}_{\text{train}}$, we construct triplets $(a, p, n)$ where $a$ is an anchor image, $p$ is a positive sampled uniformly from the same class as $a$, and $n$ is a negative sampled uniformly from a different class.
The adapter is trained with the standard triplet margin
loss:
\begin{equation}
	\mathcal{L} = \frac{1}{|\mathcal{B}|}
	\sum_{(a,p,n) \in \mathcal{B}}
	\max\!\left(0,\;
	d\!\left(\hat{\mathbf{z}}_a, \hat{\mathbf{z}}_p\right)
	- d\!\left(\hat{\mathbf{z}}_a, \hat{\mathbf{z}}_n\right)
	+ m
	\right),
	\label{eq:triplet}
\end{equation}
where $d(\cdot,\cdot)$ denotes distance, $m > 0$ is the margin, and $\hat{\mathbf{z}} = f_\theta(\mathbf{z})$ is the
transformed embedding.

A triplet $(a, p, n)$ contributes a non-zero gradient to $\theta$ if and only
if the margin constraint is \emph{violated}:
\begin{equation}
	d\!\left(\hat{\mathbf{z}}_a, \hat{\mathbf{z}}_p\right)
	- d\!\left(\hat{\mathbf{z}}_a, \hat{\mathbf{z}}_n\right)
	+ m > 0.
	\label{eq:active}
\end{equation}
We refer to such triplets as \emph{active}.
As the adapter converges on seen classes, increasingly many triplets satisfy the margin constraint and become inactive, contributing zero gradient.
This sparsity has a direct consequence for out-of-distribution generalization.
Because unseen classes are absent from $\mathcal{D}_{\text{train}}$, no triplet involving an unseen-class embedding is ever constructed.
Provided that the adapter's updates remain effectively local (i.e., the cumulative perturbation induced by seen-class training gradients remains small in unseen-class regions of the embedding space), the manifold structure for unseen classes is largely preserved.
The gradient sparsity mechanism enforces this locality: once seen-class neighborhoods satisfy the margin, the corresponding gradients vanish, and only the small fraction of actively-violated local relationships continue to be updated.


The triplet loss in Eq.~\eqref{eq:triplet} consists of hinge terms whose
subgradient with respect to $\theta$ is exactly zero whenever the margin
constraint is satisfied:
\begin{equation}
	\nabla_\theta \ell_{apn} =
	\mathbf{1}\!\left[d(\hat{\mathbf{z}}_a, \hat{\mathbf{z}}_p) -
		d(\hat{\mathbf{z}}_a, \hat{\mathbf{z}}_n) + m > 0\right]
	\cdot \nabla_\theta \delta_{apn},
	\label{eq:hinge_grad}
\end{equation}
where $\delta_{apn} = d(\hat{\mathbf{z}}_a, \hat{\mathbf{z}}_p) -
	d(\hat{\mathbf{z}}_a, \hat{\mathbf{z}}_n)$. Define the \emph{active
	triplet ratio} at training step $t$ as
\begin{equation}
	\rho(t) = \Pr_{(a,p,n)\sim\mathcal{B}}\!\left[
	d(\hat{\mathbf{z}}_a^{(t)}, \hat{\mathbf{z}}_p^{(t)}) -
	d(\hat{\mathbf{z}}_a^{(t)}, \hat{\mathbf{z}}_n^{(t)}) + m > 0
	\right].
	\label{eq:active_ratio}
\end{equation}
Under the standard assumption that training converges to a stationary point of $\mathcal{L}$, $\rho(t)$ converges to a steady-state value $\rho^* \in [0, 1]$ determined by the margin $m$ and the within-class distance distribution of the converged embeddings.
We measure $\rho^* \approx 3.5\%$ at convergence on FGVC-Aircraft (Section~\ref{sec:mechanism}), meaning that $96.5\%$ of training triplets contribute zero gradient at steady state.

Strictly speaking, sparse seen-class gradients do not formally imply that $f_\theta$ leaves unseen-class regions unchanged: the adapter is a parametric MLP, not a locally supported transformation, so any update to $\theta$ propagates globally.
The OOD measurements in Section~\ref{sec:ood} will demonstrate the evidence: EGA preserves or improves Label Precision on three unseen-class benchmarks, while ICon and SRL lack the gradient sparsity and fall below the frozen baseline.
A Lipschitz-based bound on the unseen-class perturbation is discussed in Appendix~\ref{sec:appendix_theory}.

\section{Evaluation}
\label{sec:eval}

Experiments were conducted on Chameleon Cloud~\cite{keahey2020lessons} with 256 AMD EPYC-7763 CPU cores, 512 GiB RAM, and an NVIDIA A100 GPU of 80 GB HBM2e.
Datasets include CIFAR-100 and ImageNet-1K for in-distribution, and CIFAR-10, FGVC-Aircraft, Food-101, ImageNet-1K held-out classes, and Oxford-IIIT Pet for OOD evaluation.
All datasets are available via PyTorch or Kaggle.
Backbones are CLIP ViT-B/32~\cite{clip_icml21}, DINOv2-large~\cite{oquab2024dinov}, and SigLIP~\cite{10377550}.

We compare EGA against four adapter methods: ICon~\cite{alshammariunifying}, SRL~\cite{dong2025improve}, LoRA~+~InfoNCE~\cite{hu2022lora,oord2018representation}, and LoRA~+~Triplet~\cite{hu2022lora}, all trained on a fronze backbone (CLIP unless otherwise specified) under identical splits and protocols.
ICon and SRL use capacity-matched residual adapters (${\sim}4.2$M parameters) differing from EGA only in loss, isolating the training objective.
LoRA variants use rank $r{=}128$, matching the parameter budget; we note that parameter count does not equate functional capacity (EGA permits nonlinear transformations while LoRA is low-rank), and our comparison controls the parameter budget rather than claiming architectural equivalence (see Section~\ref{sec:mechanism}).
We exclude full fine-tuning (it violates the frozen-backbone constraint required for index preservation and multi-task serving), as well as linear probing, Tip-Adapter~\cite{tipadapter_eccv22}, and prompt tuning~\cite{kzhou_ijcv22,zhou2022conditional} (incompatible with our disjoint OOD setting).
More training details and extended baseline results are in Appendices~\ref{sec:more_experiments} and~\ref{sec:more_results}, respectively.

\subsection{In-Distribution Validation}
\label{sec:indist}

\begin{figure}[t]
	\centering
	\includegraphics[width=\linewidth]{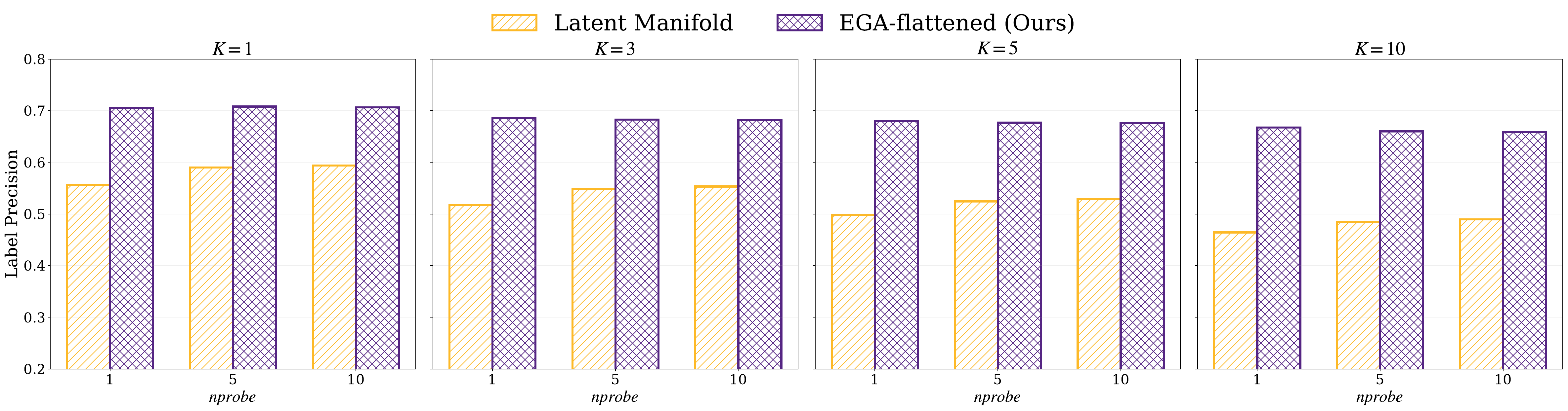}
	\caption{Label Precision@$K$ on CIFAR-100 across $K \in \{1,3,5,10\}$ and $\text{nprobe} \in \{1,5,10\}$.}
	\label{fig:precision_grid}
\end{figure}

\begin{figure}[t]
	\centering
	\includegraphics[width=\linewidth]{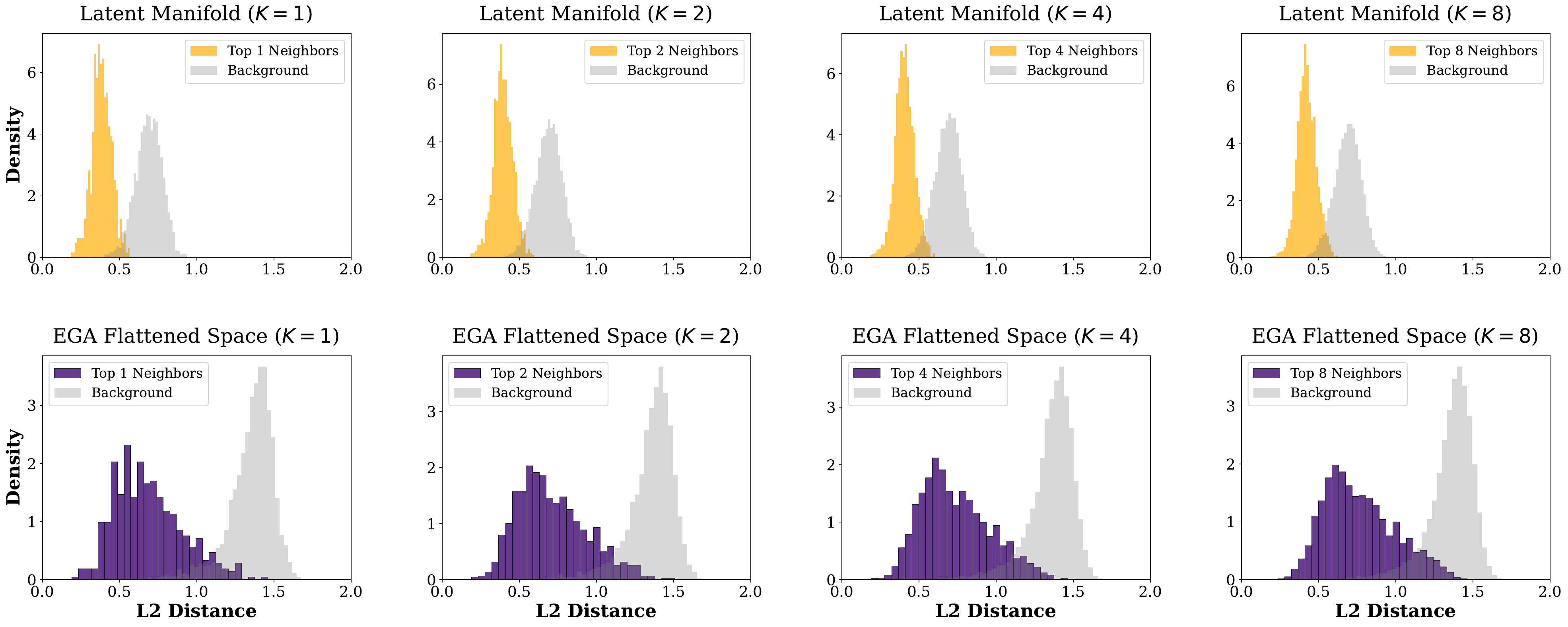}
	\caption{Distance distributions for Top-$K$ neighbors vs. random background points on CIFAR-100.}
	\label{fig:dist_grid}
\end{figure}

\begin{figure}[t]
	\centering
	\includegraphics[width=\linewidth]{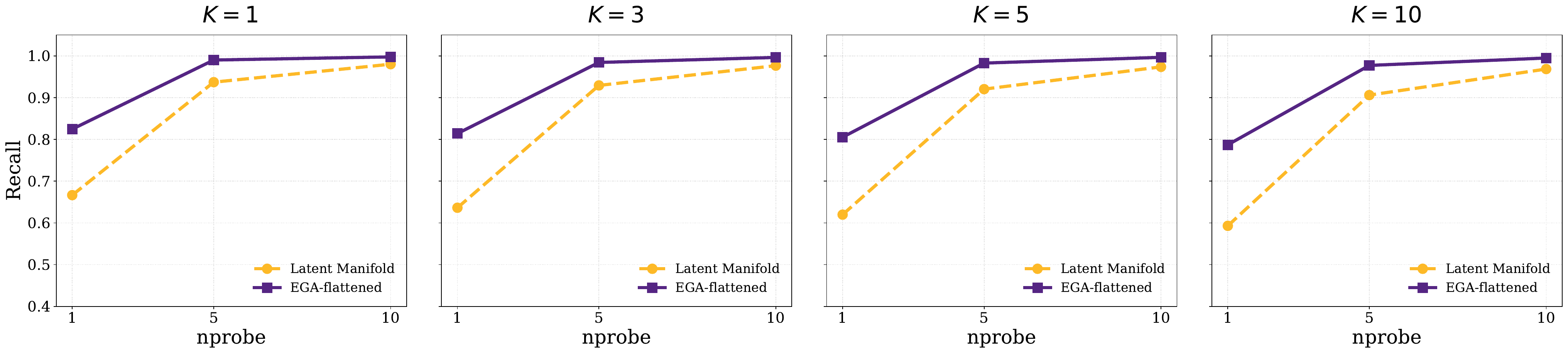}
	\caption{ANNS Recall@$K$ versus $\text{nprobe}$ on CIFAR-100.}
	\label{fig:recall_grid}
\end{figure}

Figure~\ref{fig:precision_grid} reports Label Precision@$K$ across four neighborhood sizes and three search budgets on CIFAR-100.
EGA raises Label Precision@1 from $0.549$ (raw CLIP) to $0.705$ at $\text{nprobe}=1$, and the gap persists across all $K$ and $\text{nprobe}$ values, including $\text{nprobe}=10$, indicating that the adapter tightens within-class neighborhoods rather than benefiting from broader index coverage.

The Label Precision improvement has a direct geometric correlate.
Figure~\ref{fig:dist_grid} contrasts the $L_2$ distance distributions for true Top-$K$ neighbors against random background points.
In the raw CLIP space, the two distributions overlap substantially, which is the condition under which IVF-based ANN indexing degrades.
After EGA, the two distributions separate cleanly across all $K$, and this separation drives the recall-efficiency improvement in Figure~\ref{fig:recall_grid}: raw CLIP achieves $0.667$ Recall@1 at $\text{nprobe}=1$, whereas EGA reaches $0.825$ and saturates above $0.99$ by $\text{nprobe}=5$.

\begin{wraptable}{r}{0.5\textwidth}
	\vspace{-1.5em}

	\centering
	\caption{In-distribution retrieval performance at $K=1$, $\text{nprobe}=1$ on CIFAR-100 and ImageNet-1K.
		Standard errors are below $0.01$ and omitted.
	}
	\label{tab:indist_summary}
	\small
	\setlength{\tabcolsep}{3.5pt}
	\begin{tabular}{lcccc}
		\toprule
		\multirow{2}{*}{Method} & \multicolumn{2}{c}{CIFAR-100} & \multicolumn{2}{c}{ImageNet-1K}                                      \\
		\cmidrule(lr){2-3} \cmidrule(lr){4-5}
		                        & LP@1                          & AR@1                            & LP@1           & AR@1              \\
		\midrule
		CLIP (frozen)           & 0.549                         & 0.667                           & 0.321          & 0.803             \\
		ICon                    & \textbf{0.999}                & \textbf{0.992}                  & \textbf{0.386} & \textbf{0.817}    \\
		SRL                     & \textbf{0.992}                & \textbf{0.991}                  & \textbf{0.421} & \underline{0.687} \\
		LoRA+InfoNCE            & \textbf{0.668}                & \textbf{0.879}                  & \textbf{0.426} & \textbf{0.822}    \\
		LoRA+Triplet            & \textbf{0.612}                & \textbf{0.908}                  & \textbf{0.359} & \textbf{0.870}    \\
		EGA                     & \textbf{0.705}                & \textbf{0.825}                  & \textbf{0.407} & \textbf{0.817}    \\
		\bottomrule
	\end{tabular}
	\vspace{-1em}
\end{wraptable}
Table~\ref{tab:indist_summary} reports all six methods' in-distribution performance on both CIFAR-100 and ImageNet-1K.
On CIFAR-100, ICon and SRL reach near-perfect Label Precision ($0.999$ and $0.992$) by tightening seen-class clusters under global contrastive objectives.
EGA reaches $0.705$, an improvement of $+16$ percentage points over the frozen baseline; the LoRA variants land between EGA and the baseline.
On ImageNet-1K (1000 classes), the absolute scale of all methods drops compared to CIFAR-100 because the test set contains approximately seven samples per class for retrieval.
The four trained adapters cluster within two percentage points of each other (SRL 0.421, LoRA+InfoNCE 0.426, EGA 0.407, ICon 0.386), indicating that the task is sufficiently challenging for frozen CLIP features that no lightweight adapter can substantially restructure the embedding space under a pure ID objective.
We note that ImageNet-1K as an ID task differs fundamentally from the OOD setting that motivates our work: its 1000 classes are all seen during training, and the pretrained CLIP encoder already captures rich semantic structure that adapter training can only marginally refine.
Methods that aggressively optimize ID objectives (InfoNCE, SRL) achieve slightly higher ID precision here, consistent with their design, while EGA's self-limiting triplet loss prioritizes the preservation of geometric structure that proves essential under distribution shift rather than maximizing in-distribution separation.

\subsection{Out-of-Distribution Generalization}
\label{sec:ood}

We now examine performance under strict OOD settings where evaluation categories are disjoint from training categories.
Table~\ref{tab:ood_summary} reports headline numbers at $K=1$, $\text{nprobe}=1$ across four OOD benchmarks: CIFAR-10 (trained on CIFAR-100), FGVC-Aircraft (80 seen / 20 unseen classes), Food-101 (80 seen / 21 unseen classes), and ImageNet-1K held-out classes (800 seen / 200 unseen classes).
For a fair comparison, all methods train identical adapters on top of frozen CLIP features and differ only in their loss functions.
Robustness across search budgets is reported in Appendix~\ref{sec:appendix_budget}.

\begin{table}[!t]
	\centering
	\caption{OOD performance at $K=1$, $\text{nprobe}=1$.
	}
	\label{tab:ood_summary}
	\small
	\setlength{\tabcolsep}{3pt}
	\begin{tabular}{llcccccc}
		\toprule
		Dataset             & Metric        & CLIP  & ICon                                    & SRL                                     & LoRA+InfoNCE                            & LoRA+Triplet                            & EGA                                     \\
		\midrule
		\multirow{2}{*}{CIFAR-10}
		                    & AR@1          & 0.863 & \underline{0.797}{\scriptsize$\pm$.003} & \underline{0.819}{\scriptsize$\pm$.001} & \textbf{0.869}{\scriptsize$\pm$.012}    & 0.861{\scriptsize$\pm$.011}             & \underline{0.833}{\scriptsize$\pm$.005} \\
		                    & LP@1          & 0.880 & \underline{0.560}{\scriptsize$\pm$.010} & \underline{0.536}{\scriptsize$\pm$.008} & \textbf{0.885}{\scriptsize$\pm$.006}    & \underline{0.875}{\scriptsize$\pm$.006} & \underline{0.810}{\scriptsize$\pm$.002} \\
		\midrule
		\multirow{2}{*}{Aircraft}
		                    & AR@1          & 0.773 & \textbf{0.853}{\scriptsize$\pm$.016}    & \textbf{0.879}{\scriptsize$\pm$.023}    & \textbf{0.891}{\scriptsize$\pm$.015}    & \textbf{0.893}{\scriptsize$\pm$.008}    & \textbf{0.905}{\scriptsize$\pm$.015}    \\
		                    & LP@1          & 0.512 & \underline{0.470}{\scriptsize$\pm$.034} & \underline{0.375}{\scriptsize$\pm$.032} & \textbf{0.538}{\scriptsize$\pm$.020}    & \textbf{0.569}{\scriptsize$\pm$.011}    & \textbf{0.611}{\scriptsize$\pm$.020}    \\
		\midrule
		\multirow{2}{*}{Food-101}
		                    & AR@1          & 0.842 & \underline{0.821}{\scriptsize$\pm$.001} & \underline{0.824}{\scriptsize$\pm$.015} & \textbf{0.906}{\scriptsize$\pm$.003}    & \textbf{0.893}{\scriptsize$\pm$.008}    & \textbf{0.879}{\scriptsize$\pm$.011}    \\
		                    & LP@1          & 0.881 & \underline{0.562}{\scriptsize$\pm$.011} & \underline{0.552}{\scriptsize$\pm$.010} & \underline{0.883}{\scriptsize$\pm$.004} & \underline{0.833}{\scriptsize$\pm$.007} & \underline{0.791}{\scriptsize$\pm$.002} \\
		\midrule
		\multirow{2}{*}{ImageNet-1K}
		                    & AR@1          & 0.829 & 0.849{\scriptsize$\pm$.012}             & \underline{0.747}{\scriptsize$\pm$.015} & \textbf{0.878}{\scriptsize$\pm$.008}    & \textbf{0.883}{\scriptsize$\pm$.009}    & \textbf{0.868}{\scriptsize$\pm$.011}    \\
		                    & LP@1          & 0.684 & \underline{0.620}{\scriptsize$\pm$.014} & \underline{0.665}{\scriptsize$\pm$.012} & \textbf{0.753}{\scriptsize$\pm$.009}    & \textbf{0.714}{\scriptsize$\pm$.010}    & \textbf{0.711}{\scriptsize$\pm$.008}    \\
		\midrule
		\textbf{Worst-case} & \textbf{LP@1} & 0.512 & 0.470                                   & 0.375                                   & 0.538                                   & 0.569                                   & \textbf{0.611}                          \\
		\bottomrule
	\end{tabular}
\end{table}

\paragraph{Worst-case OOD Label Precision.}
The bottom row of Table~\ref{tab:ood_summary} reports the worst-case Label Precision across the four OOD benchmarks.
EGA achieves $0.611$, exceeding LoRA+Triplet ($0.569$) by $4.2$ percentage points, LoRA+InfoNCE ($0.538$) by $7.3$ percentage points (pp), the frozen CLIP baseline ($0.512$) by $9.9$ pp, ICon ($0.470$) by $14.1$ pp, and SRL ($0.375$) by $23.6$ pp.
EGA is the only method whose Label Precision exceeds $0.6$ on every OOD benchmark we evaluate.

The four OOD benchmarks span qualitatively different shifts: CIFAR-10 is a cross-dataset shift; Aircraft and Food-101 are fine-grained intra-domain shifts where seen and unseen classes share substantial visual structure; ImageNet held-out classes is a generic large-scale shift where the pretrained encoder already provides reasonable transfer.
The worst-case reflects deployment uncertainty: the system designer commits to one adapter without knowing in advance which shift will dominate the query stream.
Methods that excel on some shifts but collapse on others (ICon collapses on Aircraft, SRL collapses on Aircraft, LoRA variants on Aircraft) cannot guarantee a quality floor.

While these four benchmarks span the major categories of distribution shift, the worst-case is driven by the fine-grained Aircraft split.
This raises the natural question of whether EGA's advantage is specific to a single dataset.
To address this, we evaluate on Oxford-IIIT Pet (37-class pet breed classification), a fine-grained benchmark from an independent visual domain (natural species rather than man-made objects), under the identical OOD protocol.
The frozen CLIP baseline already reaches LP@1 $=$ 0.9595 on unseen breeds, yet EGA further raises it to $0.9646$ while achieving the highest ANNS Recall (0.9392 vs.\ LoRA+Triplet's 0.9342; detailed numbers in Appendix~\ref{appendix:pet}).
Taken together, these two fine-grained benchmarks span both hard and easy shifts (frozen LP@1 of 0.512 on Aircraft vs.\ 0.960 on Pet), and EGA improves over the strongest baseline on both, providing evidence that its advantage is not an artifact of a single dataset or a specific difficulty scheme.

\begin{figure}[t]
	\centering
	\includegraphics[width=\linewidth]{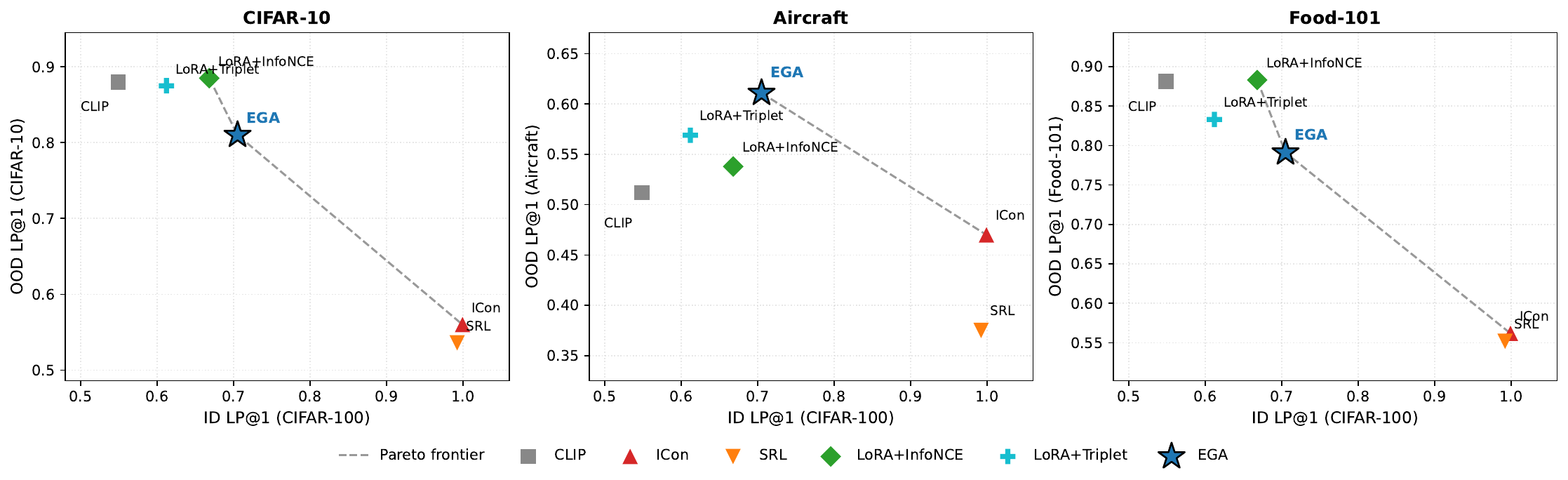}
	\caption{ID/OOD operating points of adapter training in the (ID, OOD) Label Precision plane.
	}
	\label{fig:pareto}
		\vspace{-1em}

\end{figure}

\paragraph{Per-benchmark behavior.}
On ANNS Recall, all trained adapters match or exceed the frozen baseline across Aircraft, Food-101, and ImageNet, confirming that adapter training successfully reorganizes the geometry for ANN indexing.
On Label Precision, however, the global contrastive methods fall below the frozen baseline on every fine-grained OOD dataset, with degradation exceeding 30 percentage points on Food-101 and CIFAR-10: the retrieved neighbors are geometrically closer but semantically wrong.
EGA improves Label Precision over the frozen baseline on Aircraft and ImageNet, and its degradations on CIFAR-10 and Food-101 are contained to single-digit percentage points.
The collapse of global contrastive methods is most severe exactly where unseen classes are visually similar to seen classes, consistent with our hypothesis that dense global gradients pull unseen samples into incorrect seen-class clusters.
Comparing EGA directly against LoRA+Triplet, which uses the same triplet loss but a low-rank architecture, isolates the effect of model capacity: EGA's nonlinear residual design outperforms LoRA+Triplet by $4.2$ pp on Aircraft while trailing on CIFAR-10 and Food-101 by $6.5$--$9.1$ pp, confirming that gradient sparsity and architectural capacity interact differently across shift types. 
On ImageNet, where held-out classes are visually distinct from training classes, no method collapses, and all approaches, including the frozen baseline, perform comparably.
EGA's advantage materializes precisely under difficult distribution shifts, which is the scenario where worst-case guarantees are most valuable.

EGA does not dominate every method on every individual benchmark.
Methods that restrict adapter capacity rather than the loss function achieve higher Label Precision on CIFAR-10, Food-101, and ImageNet, but suffer catastrophic degradation on the fine-grained Aircraft split, which is what determines their worst-case performance.
EGA's design explicitly prioritizes worst-case robustness; practitioners who can confidently bound the deployed distribution shift may reasonably prefer capacity-restricted alternatives.
The trade-off between these two approaches is visualized in Figure~\ref{fig:pareto}.

\subsection{Two Routes to OOD Stability: Gradient Sparsity and Capacity Restriction}
\label{sec:mechanism}


\paragraph{Route 1: Gradient sparsity (EGA).}
\begin{wrapfigure}{r}{0.59\textwidth}
	\vspace{-1em}
	\centering
	\includegraphics[width=1\linewidth]{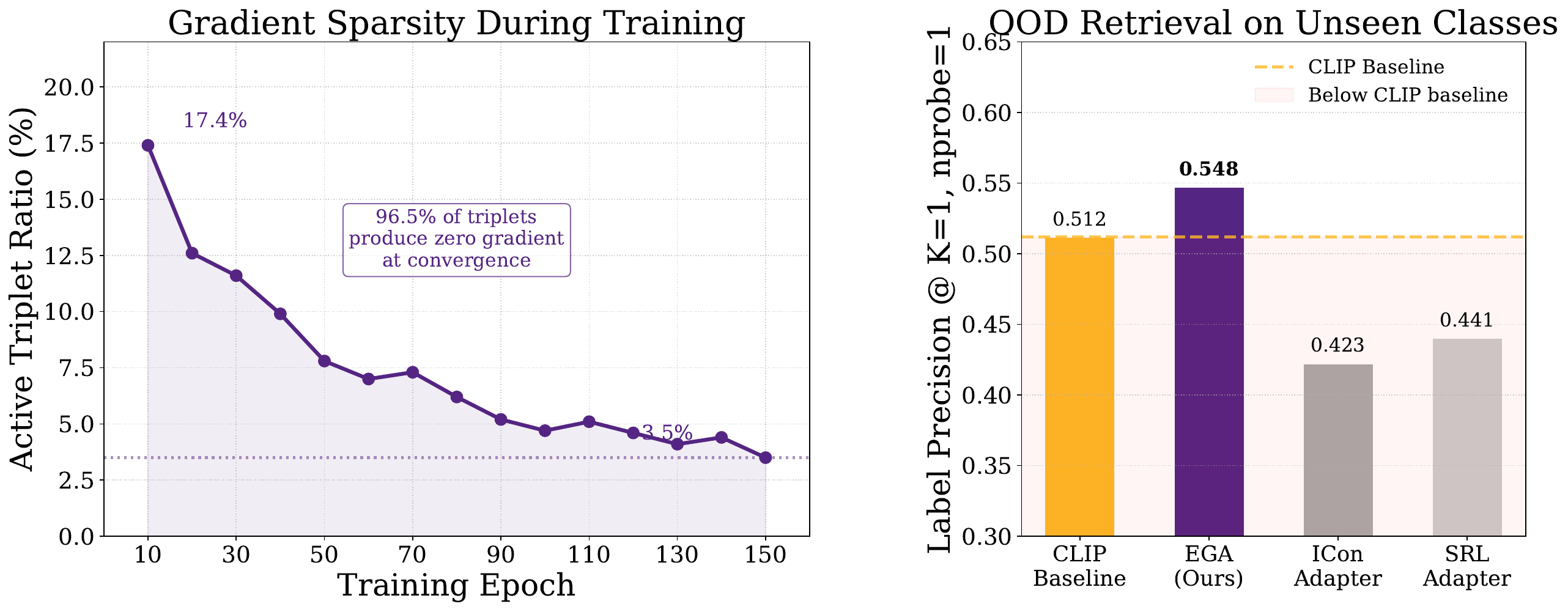}
	\caption{Gradient sparsity as OOD protection.}
	\label{fig:mechanism}
	\vspace{-1em}
\end{wrapfigure}
A triplet loss with margin $m$ produces a non-zero gradient only when $d(a, p) - d(a, n) + m > 0$.
Figure~\ref{fig:mechanism} (left) tracks the active triplet ratio from $17.4\%$ at epoch 10 to $3.5\%$ at convergence, meaning $96.5\%$ of triplets contribute zero gradient at steady state: the loss stops applying pressure once seen-class neighborhoods are locally separated, yet the adapter retains full capacity for refinement.
Sparsity neutralizes the loss-side factor while leaving capacity intact, which permits EGA's high in-distribution Label Precision (Table~\ref{tab:indist_summary}).
The self-limiting dynamic is consistent across datasets ($\rho^*$ ranges from $0.8\%$ on Food-101 to $3.5\%$ on Aircraft; Appendix~\ref{appendix:rho}), with higher residual ratios on fine-grained classes reflecting greater similarity among seen categories.
OOD protection is driven by the \emph{sparsity} of the gradient field ($96.5\%$ inactive triplets) rather than $\rho^*$.

\paragraph{Route 2: Capacity restriction (LoRA).}
The LoRA variants take the other route.
They retain a global contrastive loss (InfoNCE) or a triplet loss, but constrain the adapter to a low-rank residual update.
At rank $r=128$, LoRA's expressive power is limited to rank-$128$ perturbations of the identity, so the dense gradient field has substantially less room to reshape unseen-class geometry.
This explains why LoRA+InfoNCE, despite using the same loss family as ICon, avoids the same OOD collapse: the loss applies pressure everywhere, but the adapter cannot fully comply.
The price is paid in-distribution.
At capacity-matched scale ($\sim$4.2M parameters), LoRA+InfoNCE reaches $0.668$ Label Precision on CIFAR-100, $3.7$ percentage points below EGA, because the same low-rank constraint that bounds OOD damage also bounds in-distribution refinement.

The two routes occupy different points in Figure~\ref{fig:pareto}.
EGA, which neutralizes the loss-side factor while keeping capacity, achieves the highest worst-case OOD Label Precision.
Capacity-matched LoRA, which neutralizes the capacity-side factor, excels on three of four benchmarks but collapses on Aircraft (worst-case 0.538--0.569).
These routes are complementary: removing gradient sparsity from EGA drops Aircraft LP@1 to 0.458 (Appendix~\ref{sec:loss_arch}), while scaling LoRA rank from 64 to 512 leaves both ID and OOD nearly unchanged.
OOD robustness thus requires the combination of gradient sparsity and sufficient nonlinear capacity.

\subsection{Ablation and Sensitivity Analysis}
\label{sec:ablation}

\begin{table}[t]
		\vspace{-1em}
	\centering
	\begin{minipage}{0.38\textwidth}
		\centering
		\caption{Ablation study of EGA}
		\label{tab:ablation}
		\small
		\begin{tabular}{lc}
			\hline
			Variant                 & Accuracy \\ \hline
			Full EGA (Ours)         & 0.705    \\
			w/o Residual connection & 0.0065   \\
			w/o Zero-initialization & 0.6840   \\
			w/o L2-normalization    & 0.6590   \\ \hline
		\end{tabular}
	\end{minipage}
	\hfill
	\begin{minipage}{0.61\textwidth}
		\centering
		\caption{Sensitivity of different hyperparameters}
		\label{tab:sensitivity}
		\small
		\begin{tabular}{lcccc}
			\hline
			Triplet Margin $m$   & 0.1    & 0.2    & 0.3    & 0.5    \\ \hline
			Recall@1             & 0.8420 & 0.8180 & 0.8160 & 0.7400 \\
			Recall@5             & 0.9782 & 0.9654 & 0.9610 & 0.9125 \\ \hline
			Hidden Dimension $d$ & 512    & 1024   & 2048   & 4096   \\ \hline
			Recall@1             & 0.8300 & 0.8340 & 0.8200 & 0.8240 \\
			Recall@5             & 0.9695 & 0.9712 & 0.9680 & 0.9688 \\ \hline
		\end{tabular}
	\end{minipage}
	\vspace{-1em}
\end{table}

Table~\ref{tab:ablation} isolates the contribution of each architectural component.
The residual connection is the core element: removing it collapses accuracy from $0.705$ to $0.0065$, indicating that the adapter cannot reconstruct semantic relationships from target data alone and must operate as a refinement over the pre-trained features.
$\ell_2$-normalization keeps the transformed embeddings on the unit
hypersphere, and removing it lowers accuracy to $0.6590$.
Zero-initialization lets the adapter begin as an identity mapping, and removing it lowers accuracy to $0.6840$.

Table~\ref{tab:sensitivity} evaluates stability across hyperparameter variations.
EGA maintains consistent quality for margins $m \in \{0.1, 0.2, 0.3\}$, but increasing $m$ to $0.5$ degrades Recall@1 to $0.7400$: a larger margin forces the adapter to update embeddings that already satisfy local neighborhood constraints, reducing gradient sparsity.
Performance is also stable across hidden dimensions from 512 to 4096, indicating that EGA's behavior is determined by its structural design rather than by specific parameter choices. While the ablation shows that the residual connection is the single most critical component, the three principles are interdependent: zero-initialization ensures that the residual connection begins as an identity map, and $\ell_2$-normalization ensures that the triplet loss operates in the same metric space as the pre-trained embeddings. Their individual ablation scores understate their joint necessity.
We note that Table~\ref{tab:sensitivity} reports Recall under the in-distribution protocol of Section~\ref{sec:indist} (CIFAR-100 75/25 split); a corresponding OOD scan would require training separate adapters for each OOD benchmark at each margin value, which we leave for future work. The observed stability of ID Recall across $m \in \{0.1, 0.2, 0.3\}$ indicates that $m=0.2$ lies within a flat region of the loss landscape.

\subsection{Generalization to More Backbones}

\begin{wraptable}{r}{0.57\textwidth}
		\vspace{-1em}
    \centering
    \caption{Retrieval performance on more backbones.}
    \label{tab:ega_backbones}
    \small
    \begin{tabular}{lcccc}
        \toprule
        \multirow{2}{*}{Backbone} & \multicolumn{2}{c}{Label Precision@1} & \multicolumn{2}{c}{ANNS Recall@1} \\
        \cmidrule(lr){2-3} \cmidrule(lr){4-5}
                                  & Raw     & +EGA   & Raw     & +EGA   \\
        \midrule
        \multicolumn{5}{c}{\small In-Distribution (CIFAR-100)} \\
        CLIP ViT-B/32             & 0.549   & 0.6955 & 0.667   & 0.7985 \\
        DINOv2-large              & 0.8530  & 0.9005 & 0.8785  & 0.9310 \\
        SigLIP SO400M             & 0.7740  & 0.8595 & 0.8015  & 0.9155 \\
        \midrule
        \multicolumn{5}{c}{\small OOD (Aircraft)} \\
        CLIP ViT-B/32             & 0.512   & 0.611  & 0.773   & 0.905  \\
        DINOv2-large              & 0.6437  & 0.8593 & 0.9701  & 0.9671 \\
        SigLIP SO400M             & 0.8892  & 0.8952 & 0.9192  & 0.9760 \\
        \bottomrule
    \end{tabular}
		\vspace{-1em}
\end{wraptable}
To verify that EGA's robustness transfers to more backbones, we evaluate DINOv2-large~\cite{oquab2024dinov} and SigLIP~\cite{10377550}.
On DINOv2-large, where frozen LP@1 is $0.644$, EGA reaches $0.859$ compared to LoRA+Triplet's $0.832$, confirming that EGA's advantage persists on fine-grained shifts.
On SigLIP, where frozen LP@1 is already $0.889$, both EGA ($0.895$) and LoRA+Triplet ($0.910$) improve over the frozen baseline, with LoRA's low-rank constraint providing a slightly better inductive bias when the pretrained geometry is already strong.

The results on both backbones align with the two-routes framework of Section~\ref{sec:mechanism}: gradient sparsity is the dominant protective factor when the frozen representation is insufficient (DINOv2), while capacity restriction can offer a small advantage when the frozen geometry is already well separated (SigLIP).
Neither backbone exhibits the catastrophic OOD degradation observed with global contrastive methods on CLIP (Table~\ref{tab:ood_summary}), confirming that EGA's self-limiting dynamic transfers across backbone architectures and that gradient sparsity remains the active protective mechanism on hard shifts.
The consistency across three backbones with substantially different pre-training objectives and embedding geometries also rules out the possibility that EGA's benefits are specific to CLIP.

\section{Conclusion}
\label{sec:conclusion}

We studied adapter training for vector search under open-category queries, where the proper metric is worst-case OOD Label Precision rather than in-distribution optimality.
Global contrastive adapters silently distort unseen-class geometry, which can catastrophically degrade retrieval quality under distribution shift.
EGA addresses this through a self-limiting triplet loss: the hinge formulation automatically stops gradient flow once local class structure is satisfied, so the adapter refines only where needed while preserving the pretrained geometry elsewhere.
The residual architecture and zero initialization anchor training to the frozen representation.
A Lipschitz-based bound on the unseen-class perturbation, controlled by the empirically decaying active triplet ratio, provides analytical support for this mechanism.
Across five OOD benchmarks spanning cross-dataset, fine-grained, and large-scale shifts, EGA achieves the highest worst-case retrieval quality, with consistent improvement on two independent fine-grained domains, and transfers to more recent backbones besides CLIP.


This entire project is publicly available at \url{https://github.com/hpdic/EGA}, including all source code, paper draft, and edit history.

\begin{ack}
Results presented in this paper were obtained using the Chameleon testbed supported by the National Science Foundation.
The implementation of EGA and the preparation of this manuscript were assisted by Claude Opus 4.7 and DeepSeek V4. The ideas and analyses presented in this paper are the original work of the author, and any assistance provided by AI tools was limited to code generation and language editing under the author's supervision.
\end{ack}

\bibliography{ref}
\bibliographystyle{plain}

\clearpage
\appendix

\section{Theoretical Analysis of EGA's OOD Stability}
\label{sec:appendix_theory}

We provide a theoretical analysis to explain why the proposed EGA adapter limits the perturbation of unseen-class geometry, in contrast to global contrastive methods.
The analysis is structured in five steps, proceeding from the geometric properties of our architecture to a bound on unseen-class perturbation, and concluding with an illustrative demonstration of why sparse seen-class gradients are directionally safe for unseen categories.

\subsection{Manifold Projection and Identity Initialization}

Let the pre-trained feature space be defined as a unit hypersphere manifold $\mathcal{M} = \mathbb{S}^{d-1} \subset \mathbb{R}^d$.
For any input feature $x \in \mathcal{M}$, the residual adapter learns a continuous perturbation vector field $h_\theta: \mathcal{M} \rightarrow \mathbb{R}^d$.
The forward pass of our architecture, which combines a residual connection with $L_2$ normalization, is geometrically equivalent to a retraction mapping $\Pi_{\mathcal{M}}$ that projects the perturbed point back onto the manifold:
\begin{equation}
	f_\theta(x) = \Pi_{\mathcal{M}}(x + h_\theta(x)) = \frac{x + h_\theta(x)}{\|x + h_\theta(x)\|_2}.
\end{equation}
Under our zero initialization strategy, the initial perturbation field is strictly zero everywhere, meaning $h_{\theta_0}(x) \equiv 0$.
Consequently, the initial adapter state represents a strict identity mapping on the manifold, $f_{\theta_0}(x) = x$.
This forms the theoretical baseline for computing cumulative structural distortions: any change to an embedding is measured as a deviation from its pre-trained location.

\subsection{Path Integral of Unseen Class Distortion}

We model the training process as continuous gradient descent where parameters evolve according to $\frac{d\theta_t}{dt} = -\eta \nabla_{\theta} \mathcal{L}(\theta_t)$, with learning rate $\eta$.
For any given unseen class sample $x_u \in \mathcal{X}_{unseen}$, its total geometric drift accumulated over the training period $T$ can be formulated using the fundamental theorem of calculus as a path integral along the parameter trajectory:
\begin{equation}
	\Delta(x_u) = \left\| f_{\theta_T}(x_u) - f_{\theta_0}(x_u) \right\|_2 = \left\| \int_{0}^{T}
	J_{\theta_t}(x_u) \frac{d\theta_t}{dt} dt \right\|_2, \end{equation} where $J_{\theta_t}(x_u) = \nabla_\theta f_{\theta_t}(x_u)$ denotes the Jacobian matrix of the network evaluated at $x_u$.
Applying the triangle inequality for vector norms to the integral yields the fundamental upper bound for geometric distortion:
\begin{equation}
	\Delta(x_u) \leq \int_{0}^{T} \left\| J_{\theta_t}(x_u) \frac{d\theta_t}{dt} \right\|_2 dt.
\end{equation}

\subsection{Sparsity Modulated Deformation Bound}

We now connect the geometric distortion to the gradient sparsity of our triplet loss.
In our formulation, the parameter derivative $\frac{d\theta_t}{dt}$ is driven by the mean gradient over the mini-batch $\mathcal{B}$ of size $N$.
We assume that for any individual triplet, the gradient norm is bounded by a constant $G$, i.e., $\|\nabla_{\theta}\ell_i\|_2 \leq G$.
Therefore, the batch gradient norm satisfies:
\begin{equation}
	\left\| \nabla_{\theta} \mathcal{L}(\theta_t) \right\|_2 = \left\| \frac{1}{N} \sum_{i \in \text{active}} \nabla_{\theta} \ell_i \right\|_2 \leq \frac{1}{N} \sum_{i \in \text{active}} \left\| \nabla_{\theta} \ell_i \right\|_2 \leq \frac{\rho_t N}{N}
	G = \rho_t G, \end{equation} where $\rho_t$ is the active triplet ratio defined in Eq.~\eqref{eq:active_ratio}.
This directly bounds the parameter update rate:
\begin{equation}
	\left\| \frac{d\theta_t}{dt} \right\|_2 = \eta \left\| \nabla_{\theta} \mathcal{L}(\theta_t) \right\|_2 \leq \eta \rho_t G.
\end{equation}
To bound the integrand in Step 2, we apply the sub-multiplicativity of the spectral norm.
Assuming the network satisfies $L$-Lipschitz continuity such that the spectral norm of its Jacobian is bounded by $\left\| J_{\theta_t}(\cdot) \right\|_2 \leq L$, we obtain:
\begin{equation}
	\left\| J_{\theta_t}(x_u) \frac{d\theta_t}{dt} \right\|_2 \leq \left\| J_{\theta_t}(x_u) \right\|_2 \left\| \frac{d\theta_t}{dt} \right\|_2 \leq L (\eta \rho_t G).
\end{equation}
Substituting this back into the path integral bound yields the global perturbation upper bound for any unseen sample:
\begin{equation}
	\Delta_{EGA}(x_u) \leq \eta L G \int_{0}^{T} \rho_t dt. \label{eq:ega_bound}
\end{equation}

For methods like ICon or SRL, the loss is computed over the full pairwise similarity matrix of the batch.
In our framework, this implies that a non-zero gradient is produced for nearly all pairs in every step, corresponding to an active ratio $\rho_t^{\text{(global)}} \approx 1$.
Inserting this into Eq.~\eqref{eq:ega_bound} yields $\Delta_{\text{global}}(x_u) \leq \eta L G T$, a perturbation bound that grows linearly with training time $T$.

The critical difference is that our triplet loss with a small margin $m$ allows the optimization to reach a state where the vast majority of triplets satisfy the margin constraint.
This forces $\rho_t$ to decay to a small steady-state value $\rho^*$, as empirically observed in Section~\ref{sec:mechanism} and Appendix~\ref{appendix:rho}.
Consequently, the integral $\int_0^T \rho_t dt$ in EGA's bound is substantially smaller than $T$, keeping the cumulative perturbation limited even after prolonged training.
This analysis logically grounds the observed OOD robustness of EGA in the self-limiting nature of its training objective.

\subsection{Conjecture: Uncorrelated Residual Subspace}

The bound in Step 3 explains why the \textit{magnitude} of parameter updates is small at convergence.
However, it does not guarantee that the \textit{direction} of these residual updates is harmless to unseen classes.
We present a conjecture to address this: the small number of residual active triplets at convergence produce gradient updates whose direction is approximately uncorrelated with the Jacobian of unseen-class embeddings.

Formally, let $\mathcal{S}_{active}^{(t)} \subset \mathbb{R}^{|\theta|}$ be the tangent subspace spanned by the gradients of the active triplets at time $t$, and let $P_{\mathcal{S}^{(t)}}$ be the orthogonal projection onto it.
The parameter update $\frac{d\theta_t}{dt}$ resides entirely in $\mathcal{S}_{active}^{(t)}$, so we can write $\frac{d\theta_t}{dt} = P_{\mathcal{S}_{active}^{(t)}} \frac{d\theta_t}{dt}$.
The instantaneous perturbation on an unseen sample $x_u$ is thus $\| J_{\theta_t}(x_u) P_{\mathcal{S}_{active}^{(t)}} \frac{d\theta_t}{dt} \|_2$.

We can bound this using the sub-multiplicativity of the spectral norm:
\begin{equation}
	\left\| J_{\theta_t}(x_u) \frac{d\theta_t}{dt} \right\|_2 \leq \left\| J_{\theta_t}(x_u) P_{\mathcal{S}_{active}^{(t)}} \right\|_2 \left\| \frac{d\theta_t}{dt} \right\|_2.
\end{equation}

We conjecture that under the following conditions: (1) The distribution of unseen-class data is sufficiently diverse such that the rows of their Jacobians $J_{\theta_t}(x_u)$ are approximately isotropic in parameter space; (2) The residual active triplet subspace $\mathcal{S}_{active}^{(t)}$ at convergence is low-dimensional and its principal directions are aligned with features that discriminate between \emph{seen} classes rather than unseen classes; Then, the projection of the unseen-class Jacobian onto the active subspace, $\| J_{\theta_t}(x_u) P_{\mathcal{S}_{active}^{(t)}} \|_2$, is small with high probability over the unseen data distribution.

This conjecture is consistent with our empirical evidence: EGA preserves unseen-class geometry better than global contrastive methods (Section~\ref{sec:ood}).
However, formalizing this geometric picture would require modeling the evolution of the active subspace during training and characterizing its alignment with unseen data, a significant theoretical challenge that we leave for future work.

\subsection{Illustrative Setting: Directional Safety under Sparse Gradients.}

We illustrate the directional safety of sparse triplet gradients in a simplified setting.
Consider a frozen embedding space $\mathbb{S}^{d-1}$ with two seen classes $\mathcal{C}_1, \mathcal{C}_2$ and one unseen class $\mathcal{C}_u$.
Let $\mathbf{z}_1, \mathbf{z}_2, \mathbf{z}_u \in \mathbb{S}^{d-1}$ denote the embeddings of a query from each class after convergence of the adapter.
By construction, the adapter is trained only on triplets from $\mathcal{C}_1$ and $\mathcal{C}_2$.

A global objective (e.g., InfoNCE) encourages all pairs from the same class to be closer than pairs from different classes.
In our two-class setting, this exerts a force that pushes $\mathbf{z}_u$ toward whichever seen-class prototype $\mathbf{c}_1$ or $\mathbf{c}_2$ it happens to be nearer to in the pretrained geometry.
If $\|\mathbf{z}_u - \mathbf{c}_1\| < \|\mathbf{z}_u - \mathbf{c}_2\|$ initially, the global gradient will pull $\mathbf{z}_u$ further toward $\mathcal{C}_1$, collapsing the unseen class into the wrong seen cluster.
This is the mechanism behind the OOD degradation observed for ICon and SRL in Table~\ref{tab:ood_summary}.

Under the triplet loss, a gradient step is taken only for active triplets violating the margin.
After convergence, active triplets are sparse (Section~\ref{sec:mechanism}) and correspond almost entirely to pairs of seen classes that remain close to the decision boundary.
For an unseen sample $\mathbf{z}_u$, the instantaneous perturbation is $\Delta \mathbf{z}_u \propto \sum_{(a,p,n) \in \mathcal{A}} \nabla_{\mathbf{z}_u} \ell(a,p,n)$, where $\mathcal{A}$ is the set of active triplets.
Since no triplet involves $\mathbf{z}_u$, the gradient $\nabla_{\mathbf{z}_u} \ell$ is determined entirely by the indirect effect of parameter updates on the adapter's forward pass.

We now show that this indirect effect is small in the direction that would pull $\mathbf{z}_u$ toward a seen-class prototype.
Assume the adapter is a linear residual layer $\mathbf{z} \mapsto \ell_2(\mathbf{z} + \mathbf{W}\mathbf{z})$ with $\mathbf{W} \in \mathbb{R}^{d \times d}$ initialized to zero.
The update to $\mathbf{W}$ from an active triplet $(a,p,n)$ is a rank-1 matrix whose row space is spanned by the embeddings of $a$, $p$, and $n$.
At convergence, the active triplets are concentrated around the decision boundary between $\mathcal{C}_1$ and $\mathcal{C}_2$, so the row space of the cumulative update $\Delta\mathbf{W}$ is approximately the 2-dimensional subspace spanned by the prototypes $\mathbf{c}_1$ and $\mathbf{c}_2$.
An unseen embedding $\mathbf{z}_u$ is perturbed by $\Delta\mathbf{W}\,\mathbf{z}_u$, whose magnitude is controlled by the projection of $\mathbf{z}_u$ onto this 2-dimensional subspace.
If the unseen class is not collinear with the seen-class prototypes (i.e., $\mathbf{z}_u$ lies largely outside the span of $\mathbf{c}_1$ and $\mathbf{c}_2$), the perturbation $\|\Delta\mathbf{W}\,\mathbf{z}_u\|$ is small.
Moreover, the \emph{direction} of the perturbation is predominantly within the seen-class subspace, moving $\mathbf{z}_u$ parallel to the decision boundary rather than across it.

This illustrative analysis shows that under a linear residual adapter, sparse triplet gradients from seen classes produce parameter updates whose row space is low-dimensional and aligned with seen-class discrimination directions.
An unseen embedding whose pretrained geometry places it outside this subspace experiences only a small perturbation, and the perturbation direction is unlikely to pull it across a seen-class decision boundary.
This is in contrast to global contrastive objectives, which exert dense, radially directed forces on all embeddings regardless of whether local class structure is already satisfied.

\section{Experimental Setup Details}
\label{sec:more_experiments}

\paragraph{Backbones models.}
All experiments in the main paper use frozen CLIP ViT-B/32~\cite{clip_icml21} image embeddings as the base representation.
To verify that EGA's benefits are not specific to this backbone, we further evaluate EGA on two stronger vision encoders: DINOv2-large~\cite{oquab2024dinov} and SigLIP~\cite{10377550}.
DINOv2-large is a self-supervised vision transformer trained on 142M images, while SigLIP is a supervised vision transformer trained on 400M image-text pairs.
Both models produce higher-quality embeddings than CLIP ViT-B/32, as reflected in their stronger raw retrieval performance, and EGA continues to provide meaningful improvements on top of these stronger backbones, as shown in Table~\ref{tab:ega_backbones} in Section~\ref{sec:ablation}.

\paragraph{Adapter models.}
We compare EGA against the frozen CLIP baseline and four trained adapters, all
sharing the same training data, evaluation protocol, and 75/25 retrieval split:
\begin{itemize}
	\item \textbf{ICon}~\cite{alshammariunifying}: a global contrastive
	      objective minimizing the KL divergence between the empirical similarity
	      distribution and the class-membership distribution.
	\item \textbf{SRL}~\cite{dong2025improve}: a structural regularization
	      objective jointly optimizing batch-wide uniformity and within-class
	      homogeneity, also instantiated on the EGAMLP architecture.
	\item \textbf{LoRA~+~InfoNCE}~\cite{hu2022lora,oord2018representation}: a capacity-matched low-rank adapter (rank $r=128$, $\sim$4.2M parameters) with zero-initialized residual structure, trained with supervised InfoNCE.
	\item \textbf{LoRA~+~Triplet}~\cite{hu2022lora}: identical capacity-matched low-rank architecture as above (rank $r=128$), but trained with the same small-margin ($m=0.2$) triplet loss as EGA.
\end{itemize}

\paragraph{Datasets.}
We evaluate on six datasets covering both in-distribution and OOD retrieval scenarios.
\textbf{CIFAR-100}~(100 classes, 60K images) serves as a primary in-distribution benchmark with a 75/25 split between database and query sets.
\textbf{ImageNet-1K}~(1000 classes, 35K embedded samples) serves as a larger-scale in-distribution benchmark and additionally provides an unseen-class OOD setting via an 80/20 class split (800 seen / 200 unseen).
\textbf{CIFAR-10}~(10 classes, 60K images) provides a cross-dataset OOD setting: the adapter is trained on CIFAR-100 and evaluated on the disjoint CIFAR-10 classes.
\textbf{FGVC-Aircraft}~(100 fine-grained aircraft variants, 10K images) and \textbf{Food-101}~(101 food categories, 25K images for the test split we use) provide unseen-class OOD settings: classes are randomly partitioned into $80\%$ seen / $20\%$ unseen with a fixed seed, the adapter is trained on the seen split, and retrieval is evaluated strictly on the unseen split.
\textbf{Oxford-IIIT Pet}~(37 pet breeds, $\sim$7K images) provides an additional fine-grained OOD setting under the same 80/20 class-disjoint protocol.

\paragraph{Indexing and metrics.}
For all retrieval evaluations we use FAISS~\cite{johnson2019billion} \texttt{IndexIVFFlat} with $\text{nlist}{=}10$ centroids and report results at $\text{nprobe} \in \{1, 5, 10\}$.
We measure both Label Precision~(LP@$K$, Eq.~\eqref{eq:lp_ar}) and ANNS Recall~(AR@$K$, Eq.~\eqref{eq:lp_ar}) at $K \in \{1, 3, 5, 10\}$.

\paragraph{Training.}
All adapters are trained with AdamW (weight decay $10^{-4}$), cosine-annealed learning rate, and batch size 256 (CIFAR-100) or 512 (ImageNet-1K).
EGAMLP-based methods (EGA, ICon, SRL) use learning rate $10^{-4}$; LoRA variants use $10^{-3}$ to compensate for their smaller per-parameter scale.
Triplet-based methods sample one positive and one negative per anchor uniformly within each mini-batch; global contrastive methods (ICon, SRL, LoRA~+~InfoNCE) operate on the full pairwise similarity matrix.
All experiments use three fixed random seed (42, 123, 456) for reproducibility. For ICon and SRL, we tuned the temperature and regularization weight, respectively, on CIFAR-100 ID performance, following the protocol in their original papers; all other hyperparameters are shared across methods.

\paragraph{Compute platform.}
All experiments were conducted on the Chameleon Cloud~\cite{keahey2020lessons} platform using a compute node equipped with 256 AMD EPYC-7763 CPU cores, 512 GiB of RAM, and an NVIDIA A100 GPU of 80 GB HBM2e.
The operating system is Ubuntu 24.04 LTS.

\section{Additional Experimental Results}
\label{sec:more_results}

\subsection{Robustness across search budgets.}
\label{sec:appendix_budget}

\begin{figure}[t]
	\centering
	\includegraphics[width=\linewidth]{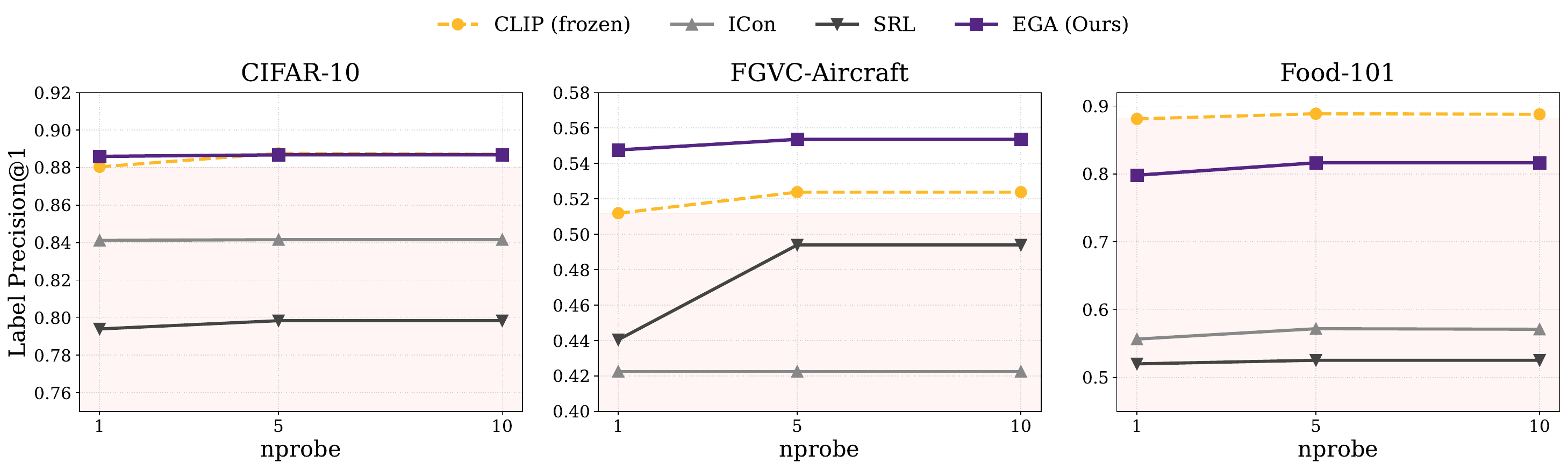}
	\caption{Label Precision@1 across three OOD benchmarks and three search budgets ($\text{nprobe} \in \{1, 5, 10\}$).}
	\label{fig:ood_lp}
\end{figure}

Figure~\ref{fig:ood_lp} traces Label Precision@1 across three OOD benchmarks and three search budgets ($\text{nprobe} \in \{1, 5, 10\}$).
The collapse pattern of ICon and SRL persists across all $18$ measurement points ($3$ datasets $\times$ $3$ budgets $\times$ $2$ methods), ruling out the possibility that the degradation is an artifact of any particular benchmark or indexing configuration.
EGA's relative ordering against other methods is also stable across budgets.

\subsection{Linear Probing Baseline Analysis}
\label{sec:linear_probing}

To address potential concerns about additional baselines, we also evaluated Linear Probing as an example.
As shown in Table~\ref{tab:linear_probing_id} and Table~\ref{tab:linear_probing_ood}, Linear Probing achieves reasonable in-distribution performance (0.5112 LP@1 on CIFAR-100) but completely collapses on out-of-distribution benchmarks (0.0065--0.0133 LP@1).
This confirms that Linear Probing is not a meaningful comparison in our cross-dataset OOD setting due to class semantic mismatch.

\begin{table}[!h]
	\centering
	\caption{Linear Probing in-distribution performance on CIFAR-100 (75/25 split, mean $\pm$ std over 3 seeds).}
	\label{tab:linear_probing_id}
	\small
	\begin{tabular}{lcc}
		\toprule
		Method         & LP@1                & AR@1                \\
		\midrule
		Linear Probing & 0.5112 $\pm$ 0.0020 & 0.7533 $\pm$ 0.0025 \\
		\bottomrule
	\end{tabular}
\end{table}

\begin{table}[!h]
	\centering
	\caption{Linear Probing OOD performance (mean $\pm$ std over 3 seeds).}
	\label{tab:linear_probing_ood}
	\small
	\begin{tabular}{lcccc}
		\toprule
		Dataset  & LP@1                & AR@1                \\
		\midrule
		Aircraft & 0.0133 $\pm$ 0.0038 & 0.4401 $\pm$ 0.0254 \\
		Food-101 & 0.0077 $\pm$ 0.0019 & 0.5162 $\pm$ 0.0228 \\
		CIFAR-10 & 0.0065 $\pm$ 0.0005 & 0.6454 $\pm$ 0.0096 \\
		\bottomrule
	\end{tabular}
\end{table}





\subsection{Disentangling Loss and Architecture}
\label{sec:loss_arch}

We conduct two controlled analyses to isolate the contributions of gradient sparsity and architecture capacity to EGA's OOD robustness.

\paragraph{Architecture without gradient sparsity.}
We train EGAMLP (the same residual architecture as EGA, $\sim$4.2M parameters) with the global contrastive InfoNCE loss under the same protocols.
Table~\ref{tab:ega_infonce} reports OOD Label Precision@1.
Under the global objective, the architecture degrades substantially across all shifts: Aircraft drops to 0.458, below even the frozen baseline (0.512), and Food-101 and CIFAR-10 lose 21.4 and 10 points relative to EGA.
This confirms that the residual design alone does not prevent unseen-class distortion; gradient sparsity from the triplet loss is essential.

\begin{table}[!h]
	\centering
	\caption{EGAMLP trained with InfoNCE on OOD benchmarks ($K{=}1$, $\text{nprobe}{=}1$).}
	\label{tab:ega_infonce}
	\small
	\begin{tabular}{lccc}
		\toprule
		OOD Benchmark & EGAMLP+InfoNCE & EGA (Triplet) & Frozen CLIP \\
		\midrule
		FGVC-Aircraft & 0.458          & 0.611         & 0.512       \\
		Food-101      & 0.667          & 0.791         & 0.881       \\
		CIFAR-10      & 0.710          & 0.810         & 0.880       \\
		\bottomrule
	\end{tabular}
\end{table}

\paragraph{Gradient sparsity with restricted capacity.}
We next examine whether increasing the capacity of a low-rank adapter can recover EGA's OOD robustness when using the triplet loss.
We train LoRA+Triplet at ranks $r \in \{64, 128, 256, 512\}$ and measure both in-distribution (CIFAR-100) and OOD (Aircraft) Label Precision@1.
As shown in Table~\ref{tab:lora_rank}, ID performance barely improves with rank (0.6456 at $r{=}64$ to 0.6480 at $r{=}512$), remaining far below EGA's 0.705, while Aircraft LP@1 never exceeds 0.566 and even deteriorates at higher ranks.
Thus, the low-rank structure intrinsically caps both ID refinement and OOD protection, regardless of parameter budget.

\begin{table}[!h]
	\centering
	\caption{LoRA+Triplet at different ranks ($K{=}1$, $\text{nprobe}{=}1$).}
	\label{tab:lora_rank}
	\small
	\begin{tabular}{lcc}
		\toprule
		LoRA Rank  & CIFAR-100 LP@1 & Aircraft LP@1 \\
		\midrule
		64         & 0.6456         & 0.5655        \\
		128        & 0.6460         & 0.5595        \\
		256        & 0.6472         & 0.5655        \\
		512        & 0.6480         & 0.5298        \\
		\midrule
		EGA (full) & 0.7050         & 0.6110        \\
		\bottomrule
	\end{tabular}
\end{table}

Together, these results establish that the combination of gradient-sparse triplet updates and a high-capacity nonlinear residual architecture is necessary for the OOD robustness observed in EGA, and that neither component alone suffices.

\subsection{Consistency of gradient sparsity across datasets.}
\label{appendix:rho}

Table~\ref{tab:rho_datasets} reports the steady-state active triplet ratio $\rho^*$ for EGA trained on three datasets.
The fraction of zero-gradient triplets exceeds $96\%$ in all cases, confirming that the self-limiting dynamic is not specific to Aircraft.
The variation in $\rho^*$ aligns with dataset difficulty: fine-grained Aircraft leaves the largest residual active fraction, while Food-101, whose classes are more visually distinct, converges to near-complete sparsity.

\begin{table}[!h]
	\centering
	\caption{Steady-state active triplet ratio $\rho^*$ across datasets.}
	\label{tab:rho_datasets}
	\begin{tabular}{lcc}
		\toprule
		Dataset       & $\rho^*$ & Zero-gradient fraction \\
		\midrule
		FGVC-Aircraft & 3.5\%    & 96.5\%                 \\
		CIFAR-100     & 2.1\%    & 97.9\%                 \\
		Food-101      & 0.8\%    & 99.2\%                 \\
		\bottomrule
	\end{tabular}
\end{table}

\subsection{Additional Fine-Grained Benchmark: Oxford-IIIT Pet}
\label{appendix:pet}

We further evaluate EGA and LoRA+Triplet on Oxford-IIIT Pet under the same 80/20 class-disjoint OOD protocol.
Table~\ref{tab:pet_ood} reports LP@1 and AR@1 at $K{=}1$, $\text{nprobe}{=}1$.
The frozen CLIP baseline achieves LP@1 $=$ 0.96 on unseen breeds, indicating that the pretrained encoder already separates these categories well; EGA further raises LP@1 to $0.9646$ while achieving the highest ANNS Recall ($0.9392$).
This demonstrates that even on easier shifts where the pretrained geometry already performs well, EGA provides a small but consistent improvement without any degradation, consistent with our finding that its advantage is most pronounced on hard fine-grained shifts.

\begin{table}[!h]
	\centering
	\caption{Oxford-IIIT Pet OOD performance ($K{=}1$, $\text{nprobe}{=}1$, 80/20 class split).}
	\label{tab:pet_ood}
	\small
	\begin{tabular}{lcc}
		\toprule
		Method       & LP@1   & AR@1   \\
		\midrule
		Frozen CLIP  & 0.9595 & 0.8785 \\
		LoRA+Triplet & 0.9519 & 0.9342 \\
		EGA          & 0.9646 & 0.9392 \\
		\bottomrule
	\end{tabular}
\end{table}

\end{document}